 \documentclass[final,3p,times]{elsarticle}

\usepackage{amssymb}
\usepackage{algorithm}
\usepackage{algorithmic}
\usepackage{setspace}
\usepackage{flushend}
\usepackage{tabularx}

\usepackage{subcaption}

\usepackage[cmex10]{amsmath}
\usepackage{color}
\usepackage{fancyhdr}

\usepackage{url}

\newcommand{\methodName}{{FRnet-DTI}}
\newcommand{\modelOne}{{FRnet-1}}
\newcommand{\modelTwo}{{FRnet-2}}

\fancypagestyle{plain}{
	\fancyhead{}
	\fancyhead[C]{first page center header}
	\fancyfoot{}
	\fancyfoot[C]{first page center footer}
}

\journal{}

\begin{document}

\begin{frontmatter}





	\title{\methodName: Deep Convolutional Neural Networks with Evolutionary and Structural Features for Drug-Target Interaction}
	
	\author[uiu]{Farshid Rayhan\corref{cor1}}
	\ead{frayhan133057@bscse.uiu.ac.bd}
	
	\author[uiu]{Sajid Ahmed}
	\author[iran]{Zaynab Mousavian}
	\author[uiu]{Dewan Md Farid}
	\author[uiu]{Swakkhar Shatabda}

	\address[uiu]{Department of Computer Science and Engineering, United International University, Bangladesh}
	\address[iran]{Department of Computer Science, University of Tehran, Iran} 
	
	\cortext[cor1]{Corresponding author}
	

\begin{abstract}
 The task of drug-target interaction prediction holds significant importance in pharmacology and therapeutic drug design. In this paper, we present {\methodName}, an auto encoder and a convolutional classifier for feature manipulation and drug target interaction prediction. Two convolutional neural neworks are proposed where one model is used for feature manipulation and the other one for classification. Using the first method {\modelOne}, we generate 4096 features for each of the instances in each of the datasets and use the second method, {\modelTwo}, to identify interaction probability employing those features. We have tested our method on four gold standard datasets exhaustively used by other researchers. Experimental results shows that our method significantly improves over the state-of-the-art method on three of the four drug-target interaction gold standard datasets on both area under curve for Receiver Operating Characteristic(auROC) and area under Precision Recall curve(auPR) metric. We also introduce twenty new potential drug-target pairs for interaction based on high prediction scores. \\
\textbf{Codes Available:} \textit{\url{https://github.com/farshidrayhanuiu/FRnet-DTI/}} \\
\textbf{Web Implementation:}  \textit{\url{http://farshidrayhan.pythonanywhere.com/FRnet-DTI/}} \\

\end{abstract}


\end{frontmatter}


\section{Introduction}
The task of drug-target interaction prediction is very important in pharmacology and therapeutic drug design. This problem can be addressed in several ways. Firstly, for a already developed drug compound the task is to find new targets with which the drug might have interactions. Secondly, for a given target protein one might search for potential drugs in the library. Another way to tackle the problem is to find the possibility of interaction given a pair of drug and target protein. In this paper, we are interested in the latter kind.  Experimental methods in predicting drug-protein interactions are expensive and time consuming and hence computational methods have been used extensively in the recent years \cite{haggarty2003multidimensional,kuruvilla2002dissecting}. 

One of the most successful computational method in drug-target interaction prediction is docking simulations \cite{kitchen2004docking}. This method largely depends on the availability of three dimensional native structure of the target protein  determined by sophisticated methods like X-Ray Crystallography. However, X-Ray Crystallography is itself a time-consuming and expensive process and thus the native structure of the targets proteins are often unavailable. These have encouraged the researchers to apply machine learning based methods to tackle the prediction problem by formulating it in a supervised learning setting \cite{mousavian2014drug}.  

Success of supervised learning methods largely depend on the training datasets. In a pioneering work on drug-target interaction prediction, Yamanishi et al. \cite{yamanishi2008prediction} proposed gold standard datasets with four sets or target proteins. Machine learning methods used in prediction of drug-target interaction often use features generated from molecular fingerprints of drugs \cite{zaynabPssm} and sequence or structure based information \cite{rayhanASFMDR17}. A good number of machine learning algorithms have been used in the literature of supervised drug-target interaction prediction that includes: Support Vector Machines (SVM) \cite{zaynabPssm}, Boosting \cite{rayhanASFMDR17}, Deep Learning \cite{chan2016large}, etc. One of the major obstacle in drug-target interaction prediction is due to the imbalance in the dataset. Since the known validated interaction among drug target pairs are not large, most of the approaches considers the unknown interactions as negative samples and thus they outnumber positive samples. The representation of the drug-target pair in the supervised learning dataset is another added challenge.

In the recent years, Chemo-genomic methods have received a lot of attention for identifying drug target interaction. They usually include methods like graph theory \cite{wang2013drug,chen2012drug}, deep learning \cite{chan2016large} , machine learning \cite{yamanishi2008prediction,bleakley2009supervised} and network analysis methods \cite{alaimo2013drug,cheng2012prediction}. In supervised learning setting, K-Nearest Neighbor(KNN) \cite{he2010predicting}, fuzzy logic \cite{xiao2013icdi}, support vector machines \cite{zaynabPssm,keum2017self} are the most commonly used classification algorithm. In \cite{yamanishi2008prediction}, drug target interaction problem was first introduced as a supervised problem and a gold standard dataset was proposed. Later those datasets have been exhaustively used by researchers. The same authors from \cite{yamanishi2008prediction}, applied distance based learning to association among pharmacological space of drug target interactions. A non-linear kernel fusion with regularized least square method was proposed by \cite{hao2016improved}.

\cite{gonen2012predicting} used Chemical and genomic kernels and Bayesian factorization. Another method , DBSI (drug based similarity interface) \cite{cheng2012prediction} proposed two dimensional chemical-structural similarity for drug similarity. Later methods like DASPfind \cite{ba2016daspfind}, NetCBP \cite{chen2013semi} , SELF-BLM \cite{keum2017self} were proposed in order to solve the problem. Bigram based features as fingerprints, extracted from position specific scoring matrix, were found very helpful solving the drug target interaction problem \cite{zaynabPssm}. Most of the supervised learning methods do not exploit the structure based features because most protein targets' three dimensional native structure are not available. \cite{huang2016systematic} used extremely randomized trees as classifier. The authors of that paper represented the drugs as molecular fingerprint and proteins as pseudo substitution matrix. Theses matrix were generated from its amino acid sequence information. Other relevant works include self-organizing theory \cite{daminelli2015common,duran2017pioneering}, similarity based methods \cite{yuan2016druge}, ensemble methods \cite{ezzat2016drug,ezzat2017drug}. A in depth literature review was done by Chen et al. on computational methods for this particular problem \cite{chen2015drug}. 

One of the most recent work was done by \cite{wen2017deep}, where the authors presented a model which consisted of multiple stacked RBM. The output layer consisted of 2 neurons each predicting the interaction and non-interaction probability respectively. \cite{chan2016large} also presented a model that used deep representations for drug target interaction predictions. In our recent work, iDTI-ESBoost \cite{rayhanASFMDR17}, we exploited evolutionary features along with structural features to predict drug protein interaction. SPIDER3, a successful secondary structural prediction tool \cite{lopez2017sucstruct,taherzadeh2017structure}, was used to generate a novel set of features for supervised learning. The novel set of features include seven primary set of features. A short description of each feature set is given in S1 of the supplementary material . In that paper, two balancing methods were used to handle the imbalance ratio of the datasets, and Adaboost \cite{freund1995desicion} was used for classification.

In this paper, we propose two deep convolutional methods for feature manipulation and predicting drug target interaction. There are named \modelOne\ and \modelTwo\ where \modelOne\ is used to generate 4096 features or deep representation of each datasets and \modelTwo\ is used for classification using the extracted features. We use the latest version of 4 gold standard datasets with 1476 features to test our method. In the experimental results using both of our method as one, we have observed magnificent auROC and auPR metric scores and therefore we strongly claim that our method is an excellent alternative for most other proposed methods for Drug-Target-Interaction (DTI).


\begin{figure*}
	\begin{center}
		\includegraphics[width=1\textwidth]{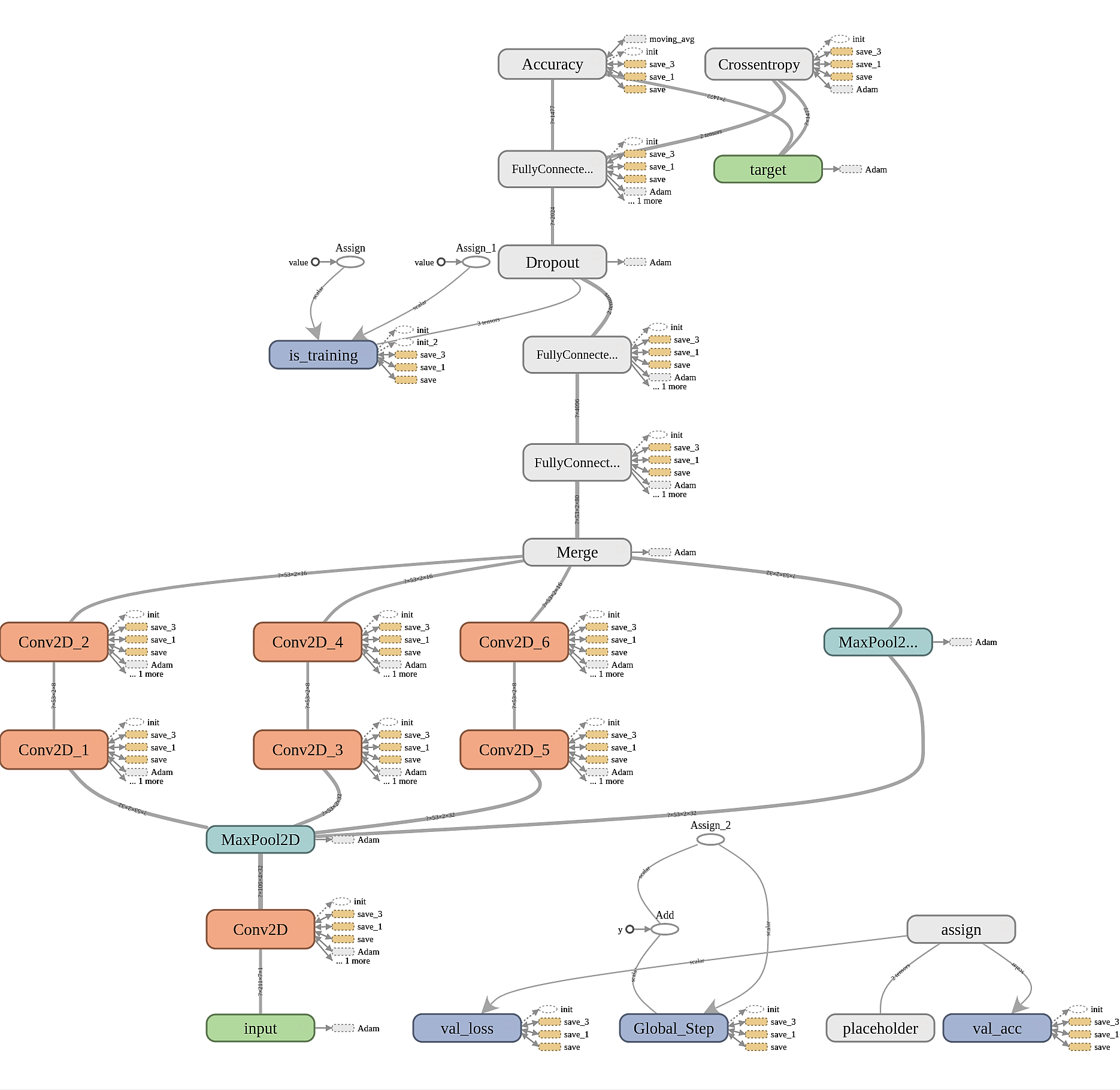}
		
		\caption{Architecture of \modelOne}\label{fig-model1}
	\end{center}
\end{figure*}

\section{Methodology}
This section provides a description of the methodology used in this paper:  algorithmic details, datasets and performance evaluation methods.

\subsection{Convolutional Models}
In this paper, we propose two novel deep learning architectures for drug target interaction, each network having its own purpose and goal. Throughout the rest of the paper these two models are referred as \modelOne\ and \modelTwo. \modelOne\ is used as a auto encoder that extracts 4096 features from the given feature sets which is than fed as input to \modelTwo\ for classification.   

The proposed models \modelOne\ and \modelTwo\ both have over millions of hyper-parameter to tune. While exhaustively tuning each of the parameter would provide the best result it is seldom done as it requires tremendous computational cost and time and risks over-fit. Four of the  most effective hyper parameters is chosen \cite{goodfellow2016practical} to tune and from a given range. Aggressive regularization is also employed to remove over-fitting to maximum extend. 

\subsection{Intuition}
Recently, Deep learning methods have been receiving a lot of attention for biological applications. In the article \cite{du2018predicting}, the authors used a model called Wide-and-Deep. Authors of \cite{wen2017deep} showed impressive improvement in prediction capability using deep learning.

The authors of \cite{wang2017computational} used stacked auto-encoders to further improve the problem of DTI. We closely follow that article and employ an auto encoder to extract features and a classifier for final classification. While the above mentioned papers address a separate problem of DTI using deep learning their intuition behind the architectural design wasn't very clear. 

We follow the architectural design of GoogleNet \cite{szegedy2014going} which is regarded as one of the most successful classification network \cite{szegedy2016rethinking, szegedy2017inception}. They use a module called the inception module (see fig: \ref{fig-model2}) where convolution operation with different filters sizes are done in parallel with a pooling operation. Each of the output are then merged together as the final output of the module. This process reliefs the burden of choosing of proper filter size or operation type between convolution and pooling. Following this intuition, we design our two models which are further explained below.

\subsubsection{\modelOne} 
In order to perform a convolution operation, a 4D tensor with the shape $[X, a, b, c]$ is required \cite{abadi2016tensorflow} where $a, b, c$ are the 3d representation of the features and $X$ is the input batch size. The latest version of the gold standard datasets were represented with 1476 features by \cite{rayhanASFMDR17} and showed very optimistic results. For convenience, a new feature with value 0 was added which extended the feature length to 1476 so that the input can be reshaped into $[X, 211, 7, 1]$. Here 211 and 7 are unique numbers and interchanging them has no effect and 1 represents that the input has only 1 channel. Basically the dataset is represented as a gray scale $211\times7$ sized image.    

\begin{figure}[h!]
			\begin{center}
				\begin{tabular}{cc}
					\includegraphics[width=0.4\textwidth]{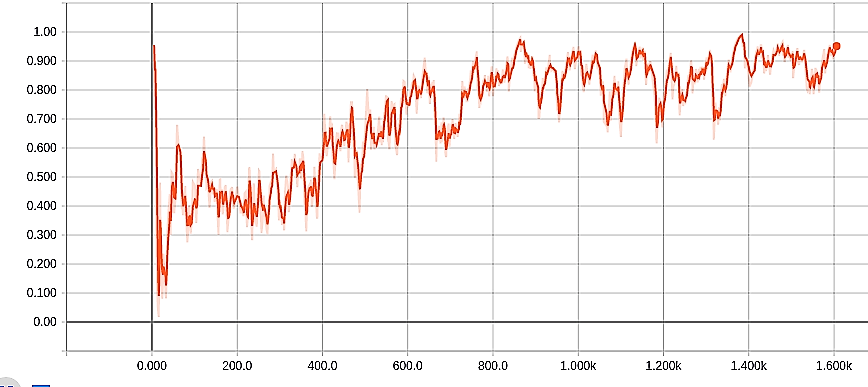}&
					\includegraphics[width=0.4\textwidth]{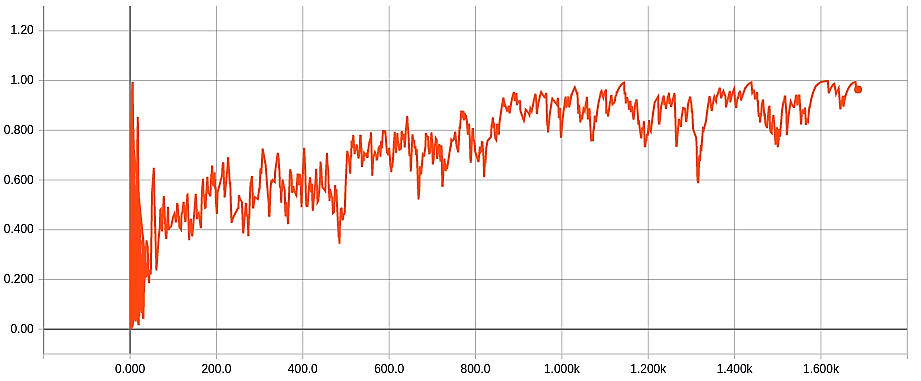} \\
					(a)&(b)\\
					\includegraphics[width=0.4\textwidth]{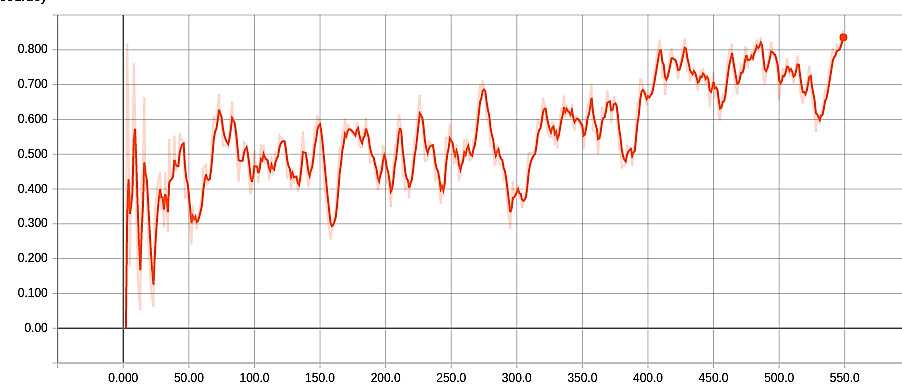}&
					\includegraphics[width=0.4\textwidth]{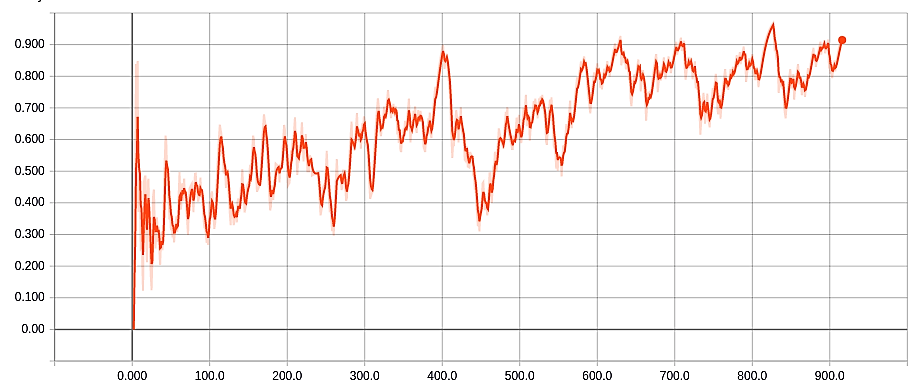} \\
					(c)&(d)\\
				\end{tabular}
			\end{center}
			
			\caption{Accuracy curves of \modelOne using  $Binary\ crossentropy$ as loss function on four datasets: (a) enzymes (b) ion channels (ic) (c) GPCRs (d) nuclear receptors (nr).\label{figCLSPR}}
		\end{figure}

This model consists of several convolutional layers, Max-Pooling layer and Fully-connected layer. Fig.~\ref{fig-model1} shows the visual representation of the complete network. Input layer takes the input in the shape of $[X, 211, 7, 1]$ and passes to a $1\times 1$ Convolutional layer with 32 filters and 2 strides which outputs a tensor with the shape $[X, 106, 4, 32]$. Here $1\times 1$ convolution means the size of the filter were $1\times 1$. `Relu' activation,  `SAME' padding and `L2' regularizer were used in each layer. After convolution, a Max-Pool operation is done with a kernel and stride value of 2 to reduce the tensor shape to $[X, 53, 2, 32]$. This network output is then fed to four parallel processes. They are depicted in Fig.~\ref{fig-model1} from left to right. The first process is a $1\times 1$ convolution with 8 filters followed by a $3\times 3$ convolution with 64 filter. Next is a $1\times 1$ convolution with 8 filters followed by a $2\times 2$ convolution with 64 filters. Then, there is a $1\times 1$ convolution with 8 filters followed by a $5\times 5$ convolution with 32 filters and lastly, a Max-pool operation with kernel and stride 1. In the next stage, a merge operator combines the 4 network outputs on the 3rd axis of the resulting tensor with the shape $[X, 53, 2, 192]$. Detailed description is provided in table \ref{shapes}. The merged network is then fed to fully connected layers with 4096 and 2048 neurons followed by a dropout operation with $Keep\_Prb$ value at $0.5$. Finally the output layer consists of 1476 neurons each neuron representing a feature value of an instance in the dataset. The model was trained with a learning rate 0.001, `$Adam$' as optimizer \cite{kingma2014adam} and $binary\_crossentropy$ as loss function. The fully connected layer with 4096 neurons is used as features to predict interaction probability using the \modelTwo\ method. The model achieved \textbf{0.85}\% accuracy just after 3 iterations and reached over 
\textbf{90}\% after 20 iteration. Accuracy curves with respect to each iteration are described in figure \ref{figCLSPR}.   

	\begin{table}[!htb]
		
		\begin{center}
			\centering
			\begin{tabular}{l|ccc}
				\hline
				Index 	& Input shape 	& Output shape after 1st ConV & Output shape after 2nd ConV/Max-pool \\ \hline
				a		&[X, 53, 2, 32]	&[X, 53, 2, 16]			&[X, 53, 2, 64]\\
				b		&[X, 53, 2, 32] &[X, 53, 2, 16]			&[X, 53, 2, 64]\\
				c		&[X, 53, 2, 32]	&[X, 53, 2, 16]			&[X, 53, 2, 32]\\
				d		&[X, 53, 2, 32] & -			            & [X, 53, 2, 32]\\ \hline
				&& Tensor shape after merge operation  & [X, 53, 2, 192] 
			\end{tabular}
		\end{center}
		\centering
		\caption{Shapes of tensor after each convolution operation leading up to the merge operation. Also know as \textbf{Inception} Operation} \label{shapes}
	\end{table}

The merge operation is used to retire the burden of choosing filter size from $1\times 1$, $2\times 2$ and $5\times 5$. In stead, the model exploits all of them and chooses the better set of features by itself. This concept was inspired form the $inception$ model \cite{szegedy2014going} which was later used to build the various versions of imageNet by the same authors \cite{szegedy2016rethinking,szegedy2017inception}. However, this model also increases computational complexity as different sized filters are used at the same time. In order to reduce some computational cost, \modelOne\ employs $1\times 1$ convolution before and after each convolution operation with different filter size to reduce computational complexity of the model. This concept was first introduced in 2013 in the article by \cite{lin2013network}. They showed that $1\times 1$ convolutional operations can be used as tool to reduce channel size of a tensor. The hypothesis of \cite{lin2013network} states that converting a tensor form shape $[X, 106, 4, 32]$ to $[X, 53, 2, 192]$ will cost much more than converting $[X, 106, 4, 32]$ to $[X, 106, 4, 16]$ using a $1\times 1$ convolution using 16 filters than converting $[X, 106, 4, 16]$ to $[X, 106, 4, 192]$ and have the same effect on the network. This same methodology is later incorporated in \modelTwo\ also.

\subsubsection{\modelTwo}
\begin{figure}[htb!]
			\begin{center}
				
				\includegraphics[width=1\textwidth]{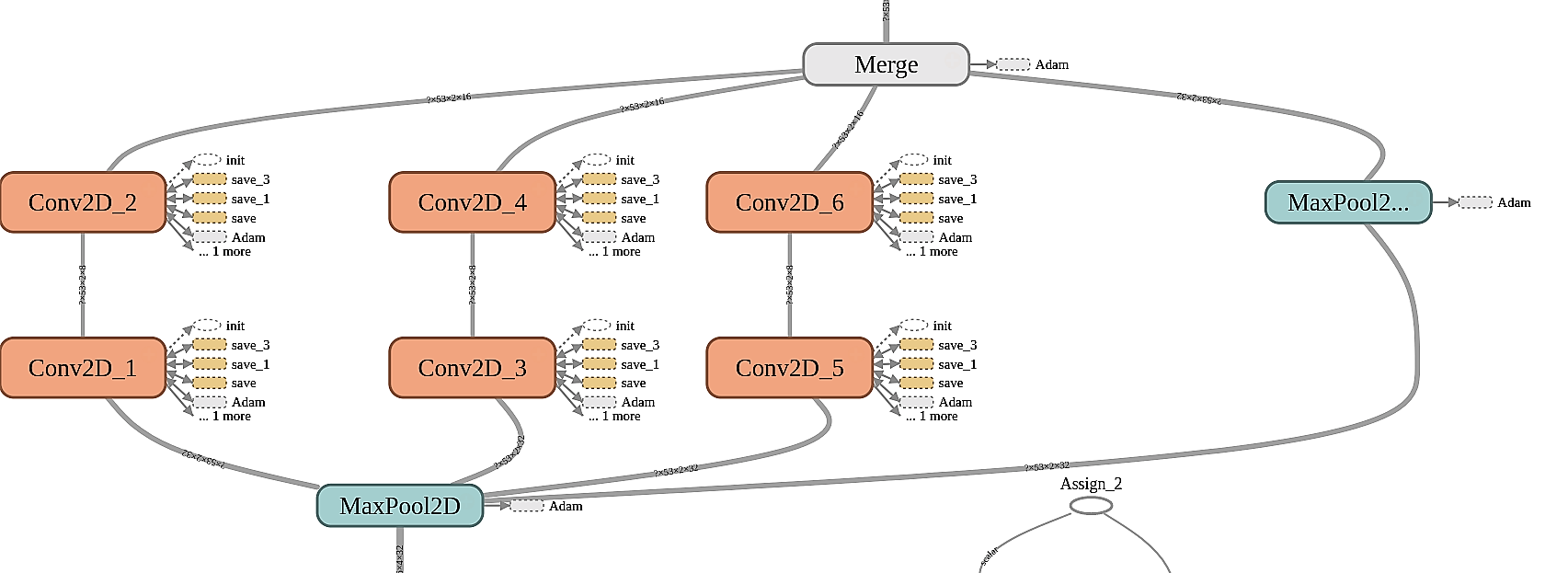}
				
			\end{center}
			
			\caption{Inception module from \cite{szegedy2016rethinking,szegedy2014going,szegedy2017inception}}\label{fig-model2}
		\end{figure}

\begin{figure*}
	\begin{center}
		\includegraphics[width=1\textwidth]{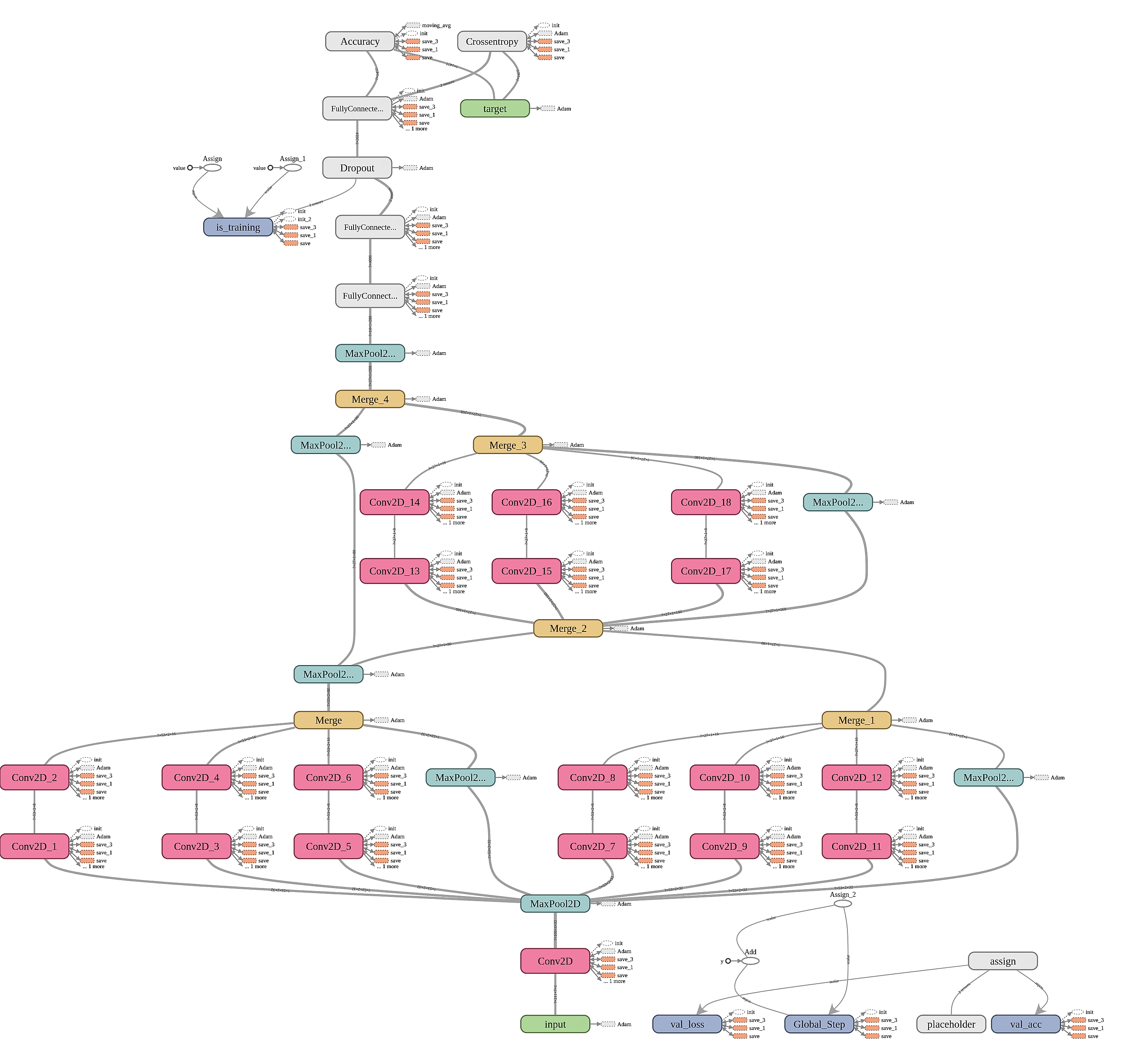}
	\end{center}
	\centering
	\caption{Architecture of \modelTwo}\label{fig-model3}
\end{figure*}

\modelTwo\ serves for the purpose of classifying interaction probability between a given drug-target pair. Similar to \modelOne\ this model employs the inception module (details figure of the model is given in figure \ref{fig-model2}). It uses the 4096 features generated by \modelOne\ as a $64\times 64\times 1$ shaped instance. In this model, the first convolution and Max-Pool operation is kept the same as the previous method. Following those operations, the tensors are parallelly fed to 2 inception modules, one with stride size of 1 (left module of Fig.~\ref{fig-model3}) and another with stride size 2.

The model merges those outputs in order to take befit of both stride size and put it through a final inception layer before connecting in to fully connected layers of 2048, 512 and finally 1 neuron for prediction. Similar to \modelOne\ this model also uses $L2$ regularization, `$Adam$' as optimizer with $Binary\_crossentropy$ as loss function with learning rate set to 0.001 and $Keep\_Prb$ value set to 0.5 in the dropout layer.      

\subsection{Datasets}

The benchmark datasets used in this article were first introduced by \cite{yamanishi2010drug} in 2010 using DrugBank \cite{wishart2008drugbank}, KEGG \cite{kanehisa2008kegg}, BRENDA \cite{schomburg2004brenda} and SuperTarget \cite{gunther2008supertarget} to extract information about drug-target interactions. These datasets are regarded as $gold\ standard$ and have been exhaustively been used by researchers throughout the years \cite{zaynabPssm,chan2016large,chen2015drug,rayhanASFMDR17}. These datasets are publicly available at: \url{http://web.kuicr.kyoto-u.ac.jp/supp/yoshi/drugtarget/}.

In this paper, an extended version of those dataset is used which consists of structural and evolutionary features. This version of dataset was first introduced in 2016 by \cite{zaynabPssm} and later further extended in 2017 by \cite{rayhanASFMDR17}. The exacted dataset from \cite{rayhanASFMDR17} were used in this project for experimentation. A short description of each dataset is given in table \ref{tabDataset}.

	\begin{table}[!htb]
		
		\begin{center}
			\begin{tabular}{l|ccccc}
				\hline
				Dataset & Drugs & Proteins & Positive Interactions & Imbalance Ratio\\
				\hline
				Enzyme			&445&664&2926&99.98\\
				Ion Chanel		&210&204&1476&28.02\\
				GPCR			&223&95&635&32.36\\
				Nuclear Receptor&54&26&90&14.6\\
				\hline
			\end{tabular}
		\end{center}
		\caption{Description of the gold standard datasets with structural and evolutionary features \cite{rayhanASFMDR17}\label{tabDataset}}
	\end{table}

\subsection{Performance Evaluation}	

A wide variety of performance metrics are available to show case and compare performance of classification models. Even though the $accuracy$ metric is sufficient enough to show the accuracy percentage of a model, in highly imbalanced dataset, such as gold standard datasets used in this experiment, that value holds little to no significance. In imbalance binary datasets, one class highly outnumbers samples of other class thus a measure of accuracy in this case makes little sense. In these type of cases, sensitivity and specificity thresholds are a very useful metric. 

Assume, P denotes the number of positive samples and N denotes the negative number of samples of a dataset. Also lets assume, $TN$ is the number of true negative and $TP$ is true positive. Similarly $FP$ represents false positive and $FN$ true negative. False positive means the negative sample that the classifier wrongly predicted as positive and false negative means a positive sample that the model has classifier predicted as negative. Conversely, True positive and true negative denotes the correctly classified positive and negative samples. 

Now from these, we can represent true positive rate or sensitivity rate as follows:
\begin{equation}
\label{Sensitivity}
Sensitivity = \frac{TP}{ TP + FN}  
\end{equation}   

Sensitivity represents the ratio of correctly predicted positive samples. Precision is the definition of the positive predictive rate (PPV). It is defined as follows:

\begin{equation}
\label{Precision}
Precision = \frac{TP}{ TP + FP}  
\end{equation}     

Precision shows precision as the percentage of accurate positive predictions of the classifiers. Specificity (SPC) or true negative rate is another important metric. It is defined as in Eq.~\ref{SPC}. False positive rate (FPR) is the representation of the ratio of the number of misclassified negative samples.

\begin{equation}
\label{SPC}
Specificity = \frac{TN}{ TN + FP}  
\end{equation}  

\begin{equation}
\label{FPR}
FPR = \frac{FP}{ TN + FP} = 1 - Specificity 
\end{equation}

Also there are two other metrics, who are independent from the imbalance ratio of the datasets, called area under curve for Receiver Operating Characteristic(auROC) and area under Precision Recall curve(auPR). Due to their ignorance towards the imbalance ratio of datasets, they have been widely used \cite{rayhanASFMDR17,chan2016large,chen2013semi,cao2012large} as standard metric for comparison. Both metrics value range from 0 to 1 where a random classifier should have a score of 0.5 and a perfect classification model will have a auPR and auROC score of 1. In both cases the higher the value the better.

Another important factor is the balance of bias and variance trade off \cite{friedman1997bias}. $k$-fold cross validation and jack knife tests are mostly used as an attempt to solve the bias-variance problem. In our experiment, we used $5$-fold shuffled cross validation on each datasets. Each time the dataset is shuffled and spitted into 5 equal parts. Then 4 of them are used for training and the rest for testing.

\section{Results and Discussion}
\label{Result_Discussion}

In the experiments reported in this paper, Python v3.6, $Tensorflow$ Library and Sci-kit learn \cite{pedregosa2011scikit} were used for the implementation. Each experiments were executed 10 times and the average result was considered. Each dataset were split into two sets, train set and test set using 5 fold cross validation. 

\begin{table}[htb!]	
	\begin{center}
			\scalebox{1}{
		\begin{tabular}{c|c|c|c}
			\hline
			Dataset  & Classifier &auPR&auROC\\
			\hline
			\hline
			enzymes		&Decision Tree	&0.28	&0.9376\\\cline{2-4}
			&SVM		&0.53	&0.9010\\	\cline{2-4}
			&MEBoost 	&0.41	&0.9404\\	\cline{2-4} 
			&CUSBoost 	&\bf0.71	&0.9345\\	\cline{2-4} 			 
			&\modelTwo\	&0.70&\bf0.9754\\	\cline{2-4}\hline
			GPCR		&Decision Tree	&0.31	&0.9038\\\cline{2-4}
			&SVM		&0.44	&0.8859\\	\cline{2-4}
			&MEBoost 	&0.46	&0.9075\\	\cline{2-4} 
			&CUSBoost 	&0.65	&0.8989\\	\cline{2-4} 			 
			&\modelTwo\	&\bf0.69&\bf0.9512\\	\cline{2-4}\hline
			Ion Channel	&Decision Tree	&0.29	&0.933\\\cline{2-4}
			&SVM		&0.40	&0.8904\\	\cline{2-4}
			&MEBoost 	&0.39	&0.928\\	\cline{2-4} 
			&CUSBoost 	&0.45	&0.8851\\	\cline{2-4} 			 
			&\modelTwo\	&\bf0.49&\bf0.9478\\	\cline{2-4}\hline
			NR			&Decision Tree	&0.46	&0.8147\\\cline{2-4}
			&SVM		&0.41	&0.7605\\	\cline{2-4} 
			&MEBoost 	&0.23	&0.9165\\	\cline{2-4} 
			&CUSBoost 	&0.71	&0.8989\\	\cline{2-4} 
			&\modelTwo\		&\bf0.73&\bf0.9241\\	\cline{2-4}\hline

			\hline
			
		\end{tabular}}
	\end{center}
	\caption{A comparison of performances among \modelTwo\ and other classifiers on the gold standard datasets in terms of auROC and auPR using 4096 features generated by \modelOne. \label{tabWeakClassifiers}}
\end{table}

\modelOne\ method is multilayer deep auto-encoder that uses convolution, max-pool and fully connected layers to regenerate the input as output in the final fully connected layer. For each of the datasets, the model was trained to achieve accuracy over of \textbf{95}\%. Due to the use of aggressive regularization, $Keep\_Prb$ value of 0.5 in dropout and $L2$ regularization in each layer using learning rate of 0.001  to avoid over fitting, the models were unable to achieve accuracy higher than \textbf{97}\% for any of the datasets. The first fully connected layer in the network has 4096 neurons in the output and those were used to extract 4096 features from each dataset. For a fair sake of comparison, \modelTwo\ was tested with several state of the art machine learning algorithms like, Decision Tree \cite{safavian1991survey}, SVM \cite{joachims1998making}, MEBoost \cite{rayhan2017cusboost} and CUSBoost \cite{rayhan2017meboost}. Each of these classifiers were fed the 4096 features generated by \modelOne. Results in terms of auPR and auROC are given Table~\ref{tabWeakClassifiers}. Table~\ref{tabWeakClassifiers} shows that \modelTwo\ is able to produce results with better auROC for all the datasets. However, in terms of auPR the results in three datasets are better than the competitor algorithms. However, for the `enzymes' dataset, the performance of \modelTwo\ is very close to the best performing CUSBoost. Note that other classifiers also achieved very impressive auROC and auPR score which shows the effectiveness of the features generated by \modelOne. Therefore even though \modelOne\ is designed for feature manipulation of the four datasets mentioned in this article, it can be used as a strong feature manipulation tool on other domains as well.

\begin{table}
	\begin{center}
					\scalebox{1}{
			\begin{tabular}{c|c|c|c|c}
				\hline
				Dataset & Reference & Classifier &auPR&auROC\\
				\hline
				\hline
				enzymes	&\cite{rayhanASFMDR17}		&AdaBoost		&0.68&0.9689\\\cline{3-5}
				&\cite{rayhanASFMDR17}		&Random Forest	&0.43	&0.9457\\\cline{3-5}
				&\cite{zaynabPssm}			&SVM			&0.54	&0.9194\\	\cline{3-5} 
				&	&\modelTwo		&\bf0.70&\bf0.9754\\	\cline{2-5}\hline
				GPCR	&\cite{rayhanASFMDR17}		&AdaBoost		&0.31	&0.9128\\\cline{3-5}
				&\cite{rayhanASFMDR17}		&Random Forest	&0.30	&0.9168 \\\cline{3-5}
				&\cite{zaynabPssm}			&SVM			&0.28	&0.8720\\	\cline{3-5} 
				&&\modelTwo 								&\bf0.69&\bf0.9512\\	\cline{2-5}\hline
				Ion Channel	&\cite{rayhanASFMDR17}		&AdaBoost	&0.48	&0.9369\\\cline{3-5}
				&\cite{rayhanASFMDR17}		&Random Forest	&0.40	&0.9234 \\\cline{3-5}
				&\cite{zaynabPssm}			&SVM			&0.39	&0.8890\\	\cline{3-5} 
				&&\modelTwo									&\bf0.49&\bf0.9512\\	\cline{2-5}\hline
				NR			&\cite{rayhanASFMDR17}		&AdaBoost	&\bf0.79	&\bf0.9285\\\cline{3-5}
				&\cite{rayhanASFMDR17}		&Random Forest	&0.29	&0.7723 \\\cline{3-5}
				&\cite{zaynabPssm}			&SVM			&0.41	&0.8690\\	\cline{3-5} 
				&							& \modelTwo		&0.73	&0.9241\\	\cline{2-5}\hline

				\hline
				
			\end{tabular}}
	\end{center}
	\caption{A performance comparison among \modelTwo\ with AdaBoost, Support Vector Machine and Random Forest classifiers on the gold standard datasets auROC and auPR curve \label{tabClassifiers}}
\end{table}      

We have compared our method with other state of the art classifiers mentioned in recent literatures such as SVM, AdaBoost and random forest. \modelTwo\ shows superior performance in both metric on all the datasets except for NR which holds only 1048 instances and is the smallest dataset among the others. Comparison with other classification models are shown in Table~\ref{tabClassifiers}. Results for the other methods were taken from the experiments reported in the literature \cite{rayhanASFMDR17,zaynabPssm}. Note that, for each of the datasets except the nuclear receptor (NR) dataset, performance of \modelTwo\ is superior to the other methods both in terms of auPR and auROC. For the NR dataset, the performance of \modelTwo\ is almost similar to the best performing boosting classifier. The auPR value is second best and probably because of the fact that this dataset is highly clustered and clustered sampling techniques for balancing used in \cite{rayhanASFMDR17} makes it perform better in this particular case.

\begin{table}
	\begin{center}
				\scalebox{1}{
			\begin{tabular}{l|c|c|c|c}
				\hline
				Dataset &Enzyme	&GPCR &ion channels	&nuclear receptor \\ \hline
				\cite{yamanishi2008prediction}		& 0.904 & 0.8510 & 0.8990 & 0.8430 \\
				\cite{yamanishi2010drug}		& 0.8920 & 0.8120 & 0.8270 & 0.8350 \\
				\cite{cheng2012prediction}		&	0.8075	&	0.8029			&	0.8022	&  0.7578\\	
				\cite{gonen2012predicting}		& 0.8320 & 0.7990 & 0.8570 & 0.8240 \\
				\cite{chen2013semi}		& 0.8251 & .8034 & 0.8235 & 0.8394 \\
				\cite{wang2013drug}		& 0.8860 & 0.8930 & 0.8730 & 0.8240 \\
				\cite{mutowo2016drug}		& 0.9480 & 0.8990 & 0.8720 & 0.8690 \\
				\cite{rayhanASFMDR17} 		& 0.9689 & 0.9369 & 0.9222 & \bf0.9285 \\
				Our Method					& \bf0.9754 & \bf0.9478 & \bf0.9512 & 0.9241    \\
				
				\hline
			\end{tabular}}
	\end{center}
	\caption{Performance of \modelTwo\ on the four benchmark gold datasets in terms of auROC with comparison to other state-of-the-art methods.\label{tabMain}}
\end{table} 

We have also compared our results against methods which used unsupervised and semi-supervised methods reported in the literature \cite{cheng2012prediction,gonen2012predicting, chen2013semi, yamanishi2008prediction,yamanishi2010drug,wang2013drug}. Table~\ref{tabMain} shows a comparisons of auROC scores of other methods including supervised methods \cite{zaynabPssm,rayhanASFMDR17}. Note that, our proposed method achieves significantly higher auROC for three datasets among four and for the NR dataset, the performance is only second best and very close to the best performing one.

\begin{table}
	
	\begin{center}
		\scalebox{1}{
			\begin{tabular}{l|c|c|c|c}
				\hline
				Predictor & enzymes &  GPCRs & ion channels & nuclear receptors \\\hline
				\cite{zaynabPssm} &0.54&0.39&0.28&0.41\\
				\cite{rayhanASFMDR17} & 0.68& 0.48&0.48&\bf 0.79\\
				\modelTwo &\bf 0.70&\bf 0.69&\bf 0.49& 0.73\\\hline
			\end{tabular}
		}
	\end{center}
	\caption{Comparison of the performance of \modelTwo\ on the four benchmark gold datasets from \cite{rayhanASFMDR17} in terms of auPR with other the state-of-the-art methods.\label{tabMainPR}}
\end{table}

In the literature of imbalanced classification problems, it has been often argued that between area under Precision Recall curve (auPR) and area under Receiver Operating Curve (auROC), auPR should be considered more significant. However, only \cite{zaynabPssm} and \cite{rayhanASFMDR17} reported auPR scores in their paper. A comparison in terms of auPR score are given in Table~\ref{tabMainPR}. Here too, its interesting to note the superior performance of our proposed model on all the datasets.

We have also tested our method with input shape $[X, 7, 211, 1]$ instead of $[X, 211 , 7, 1]$ and found similar results which concludes that changing the input shape has little to no effect on the performance on the models. Results using input shape $[X, 7, 211, 1]$ is provided in table \ref{tabWeakClassifiers2}.

			\begin{table}[!htb]
				
				\begin{center}
								\scalebox{1}{
					\begin{tabular}{c|c|c|c}
						\hline
						Dataset  & Classifier &auPR&auROC\\
						\hline
						\hline
						enzymes			&Decision Tree	&0.27	&0.9299\\\cline{2-4}
						&SVM			&0.54	&0.9035\\	\cline{2-4} 
						&\modelTwo\		&\bf0.70&\bf0.9713\\	\cline{2-4}\hline
						GPCR			&Decision Tree	&0.32	&0.9038\\\cline{2-4}
						&SVM			&0.48	&0.8859\\	\cline{2-4} 
						&\modelTwo\		&\bf0.70&\bf0.9255\\	\cline{2-4}\hline
						Ion Channel		&Decision Tree	&0.60	&0.9235\\\cline{2-4}
						&SVM			&0.52	&0.8894\\	\cline{2-4} 
						&\modelTwo\		&\bf0.50&\bf0.9507\\	\cline{2-4}\hline
						NR				&Decision Tree	&0.43	&0.8207\\\cline{2-4}
						&SVM			&0.42	&0.7588\\	\cline{2-4} 
						&\modelTwo\	&\bf0.62&\bf0.9134\\\cline{2-4}\hline

						\hline
						
					\end{tabular}}
				\end{center}
				\caption{Performance comparison of \modelTwo\ and other classifiers on the gold standard datasets in terms of auROC and auPR using 4096 features generated by \modelOne\ with input shape [X, 7, 211, 1]. \label{tabWeakClassifiers2}}
			\end{table}

Since the prediction scores with high confidence are interesting in practical applications, A list of top five the false positive interactions based on \modelTwo's prediction score is given in S2 of the supplementary materials. These are the interactions that are known as not interacting pair but the model highly suggests other wise.

\section{Conclusion}
\label{Conclusion}
In This paper, we propose two novel deep neural net architectures, \modelOne\ and \modelTwo\ where \modelOne\ aims to extract convolutional features and \modelTwo\ tries to identify drug target interaction using the extracted features. From \cite{rayhanASFMDR17}, we exploit our algorithm with datasets consisting with both structural and evolutionary features and with the help of \modelOne\ we try to generate 4096 informative features. These datasets are regarded as gold standard datasets and are exhaustively used by researchers. We have conducted extensive experiments and produced the results in term of auROC and auPR scores. In many previous literatures like \cite{zaynabPssm,rayhanASFMDR17}, it was argued that in case of drug target interaction, it is more appropriate to use auPR metric over auROC as the gold standard datasets are highly imbalanced with very few interaction samples. For this reason, \modelTwo\ was focused on getting a superior auPR score even by sacrificing auROC score. We also proposed 5 new possible interaction pair for each of the 4 datasets based on prediction score. Up to this moment, our proposed method outperforms other state of that art methods in 3 of the 4 benchmark datasets in auPR and auROC metric. We believe the excellent performance our method will motivate other practitioners and researchers to exploit both methods for not only drug target interaction but also in other domains.      

\section*{Author Contributions}
FR and SS initiated the project with the idea of using structural features. FR proposed and implemented the convolutional architecture under supervision of SS. All the methods, algorithms and results have been analyzed and verified by MSR and the other 2 authors. MSR and SS provided significant biological insights. FR prepared the manuscript with help from SS where the other authors contributed in the process and approved the final version.

\section*{Competing Interest}

The authors declare that they have no competing interests. \\


\bibliographystyle{elsarticle-harv}
\bibliography{drugtarget}

\begin{thebibliography}{53}
\expandafter\ifx\csname natexlab\endcsname\relax\def\natexlab#1{#1}\fi
\expandafter\ifx\csname url\endcsname\relax
  \def\url#1{\texttt{#1}}\fi
\expandafter\ifx\csname urlprefix\endcsname\relax\def\urlprefix{URL }\fi

\bibitem[{Abadi et~al.(2016)Abadi, Barham, Chen, Chen, Davis, Dean, Devin,
  Ghemawat, Irving, Isard, et~al.}]{abadi2016tensorflow}
Abadi, M., Barham, P., Chen, J., Chen, Z., Davis, A., Dean, J., Devin, M.,
  Ghemawat, S., Irving, G., Isard, M., et~al., 2016. Tensorflow: A system for
  large-scale machine learning. In: OSDI. Vol.~16. pp. 265--283.

\bibitem[{Alaimo et~al.(2013)Alaimo, Pulvirenti, Giugno, and
  Ferro}]{alaimo2013drug}
Alaimo, S., Pulvirenti, A., Giugno, R., Ferro, A., 2013. Drug--target
  interaction prediction through domain-tuned network-based inference.
  Bioinformatics 29~(16), 2004--2008.

\bibitem[{Ba-Alawi et~al.(2016)Ba-Alawi, Soufan, Essack, Kalnis, and
  Bajic}]{ba2016daspfind}
Ba-Alawi, W., Soufan, O., Essack, M., Kalnis, P., Bajic, V.~B., 2016. Daspfind:
  new efficient method to predict drug--target interactions. Journal of
  cheminformatics 8~(1), 15.

\bibitem[{Bleakley and Yamanishi(2009)}]{bleakley2009supervised}
Bleakley, K., Yamanishi, Y., 2009. Supervised prediction of drug--target
  interactions using bipartite local models. Bioinformatics 25~(18),
  2397--2403.

\bibitem[{Cao et~al.(2012)Cao, Liu, Xu, Lu, Huang, Hu, and
  Liang}]{cao2012large}
Cao, D.-S., Liu, S., Xu, Q.-S., Lu, H.-M., Huang, J.-H., Hu, Q.-N., Liang,
  Y.-Z., 2012. Large-scale prediction of drug--target interactions using
  protein sequences and drug topological structures. Analytica chimica acta
  752, 1--10.

\bibitem[{Chan et~al.(2016)Chan, You, et~al.}]{chan2016large}
Chan, K.~C., You, Z.-H., et~al., 2016. Large-scale prediction of drug-target
  interactions from deep representations. In: Neural Networks (IJCNN), 2016
  International Joint Conference on. IEEE, IEEE, pp. 1236--1243.

\bibitem[{Chen and Zhang(2013)}]{chen2013semi}
Chen, H., Zhang, Z., 2013. A semi-supervised method for drug-target interaction
  prediction with consistency in networks. PloS one 8~(5), e62975.

\bibitem[{Chen et~al.(2012)Chen, Liu, and Yan}]{chen2012drug}
Chen, X., Liu, M.-X., Yan, G.-Y., 2012. Drug--target interaction prediction by
  random walk on the heterogeneous network. Molecular BioSystems 8~(7),
  1970--1978.

\bibitem[{Chen et~al.(2015)Chen, Yan, Zhang, Zhang, Dai, Yin, and
  Zhang}]{chen2015drug}
Chen, X., Yan, C.~C., Zhang, X., Zhang, X., Dai, F., Yin, J., Zhang, Y., 2015.
  Drug--target interaction prediction: databases, web servers and computational
  models. Briefings in bioinformatics 17~(4), 696--712.

\bibitem[{Cheng et~al.(2012)Cheng, Liu, Jiang, Lu, Li, Liu, Zhou, Huang, and
  Tang}]{cheng2012prediction}
Cheng, F., Liu, C., Jiang, J., Lu, W., Li, W., Liu, G., Zhou, W., Huang, J.,
  Tang, Y., 2012. Prediction of drug-target interactions and drug repositioning
  via network-based inference. PLoS Comput Biol 8~(5), e1002503.

\bibitem[{Daminelli et~al.(2015)Daminelli, Thomas, Dur{\'a}n, and
  Cannistraci}]{daminelli2015common}
Daminelli, S., Thomas, J.~M., Dur{\'a}n, C., Cannistraci, C.~V., 2015. Common
  neighbours and the local-community-paradigm for topological link prediction
  in bipartite networks. New Journal of Physics 17~(11), 113037.

\bibitem[{Du et~al.(2018)Du, Wang, Wang, Chen, and Chang}]{du2018predicting}
Du, Y., Wang, J., Wang, X., Chen, J., Chang, H., 2018. Predicting drug-target
  interaction via wide and deep learning. In: Proceedings of the 2018 6th
  International Conference on Bioinformatics and Computational Biology. ACM,
  pp. 128--132.

\bibitem[{Dur{\'a}n et~al.(2017)Dur{\'a}n, Daminelli, Thomas, Haupt, Schroeder,
  and Cannistraci}]{duran2017pioneering}
Dur{\'a}n, C., Daminelli, S., Thomas, J.~M., Haupt, V.~J., Schroeder, M.,
  Cannistraci, C.~V., 2017. Pioneering topological methods for network-based
  drug--target prediction by exploiting a brain-network self-organization
  theory. Briefings in Bioinformatics, bbx041.

\bibitem[{Ezzat et~al.(2016)Ezzat, Wu, Li, and Kwoh}]{ezzat2016drug}
Ezzat, A., Wu, M., Li, X.-L., Kwoh, C.-K., 2016. Drug-target interaction
  prediction via class imbalance-aware ensemble learning. BMC bioinformatics
  17~(19), 509.

\bibitem[{Ezzat et~al.(2017)Ezzat, Wu, Li, and Kwoh}]{ezzat2017drug}
Ezzat, A., Wu, M., Li, X.-L., Kwoh, C.-K., 2017. Drug-target interaction
  prediction using ensemble learning and dimensionality reduction. Methods.

\bibitem[{Freund and Schapire(1995)}]{freund1995desicion}
Freund, Y., Schapire, R.~E., 1995. A desicion-theoretic generalization of
  on-line learning and an application to boosting. In: European conference on
  computational learning theory. Springer, Springer, pp. 23--37.

\bibitem[{Friedman(1997)}]{friedman1997bias}
Friedman, J.~H., 1997. On bias, variance, 0/1—loss, and the
  curse-of-dimensionality. Data mining and knowledge discovery 1~(1), 55--77.

\bibitem[{G{\"o}nen(2012)}]{gonen2012predicting}
G{\"o}nen, M., 2012. Predicting drug--target interactions from chemical and
  genomic kernels using bayesian matrix factorization. Bioinformatics 28~(18),
  2304--2310.

\bibitem[{Goodfellow et~al.(2016)Goodfellow, Bengio, and
  Courville}]{goodfellow2016practical}
Goodfellow, I., Bengio, Y., Courville, A., 2016. Practical methodology. Deep
  Learning.

\bibitem[{G{\"u}nther et~al.(2008)G{\"u}nther, Kuhn, Dunkel, Campillos, Senger,
  Petsalaki, Ahmed, Urdiales, Gewiess, Jensen, et~al.}]{gunther2008supertarget}
G{\"u}nther, S., Kuhn, M., Dunkel, M., Campillos, M., Senger, C., Petsalaki,
  E., Ahmed, J., Urdiales, E.~G., Gewiess, A., Jensen, L.~J., et~al., 2008.
  Supertarget and matador: resources for exploring drug-target relationships.
  Nucleic acids research 36~(suppl 1), D919--D922.

\bibitem[{Haggarty et~al.(2003)Haggarty, Koeller, Wong, Butcher, and
  Schreiber}]{haggarty2003multidimensional}
Haggarty, S.~J., Koeller, K.~M., Wong, J.~C., Butcher, R.~A., Schreiber, S.~L.,
  2003. Multidimensional chemical genetic analysis of diversity-oriented
  synthesis-derived deacetylase inhibitors using cell-based assays. Chemistry
  \& biology 10~(5), 383--396.

\bibitem[{Hao et~al.(2016)Hao, Wang, and Bryant}]{hao2016improved}
Hao, M., Wang, Y., Bryant, S.~H., 2016. Improved prediction of drug-target
  interactions using regularized least squares integrating with kernel fusion
  technique. Analytica chimica acta 909, 41--50.

\bibitem[{He et~al.(2010)He, Zhang, Shi, Hu, Kong, Cai, and
  Chou}]{he2010predicting}
He, Z., Zhang, J., Shi, X.-H., Hu, L.-L., Kong, X., Cai, Y.-D., Chou, K.-C.,
  2010. Predicting drug-target interaction networks based on functional groups
  and biological features. PloS one 5~(3), e9603.

\bibitem[{Huang et~al.(2016)Huang, You, and Chen}]{huang2016systematic}
Huang, Y.-A., You, Z.-H., Chen, X., 2016. A systematic prediction of
  drug-target interactions using molecular fingerprints and protein sequences.
  Current protein \& peptide science.

\bibitem[{Joachims(1998)}]{joachims1998making}
Joachims, T., 1998. Making large-scale svm learning practical. Tech. rep.,
  Technical report, SFB 475: Komplexit{\"a}tsreduktion in Multivariaten
  Datenstrukturen, Universit{\"a}t Dortmund.

\bibitem[{Kanehisa et~al.(2008)Kanehisa, Araki, Goto, Hattori, Hirakawa, Itoh,
  Katayama, Kawashima, Okuda, Tokimatsu, et~al.}]{kanehisa2008kegg}
Kanehisa, M., Araki, M., Goto, S., Hattori, M., Hirakawa, M., Itoh, M.,
  Katayama, T., Kawashima, S., Okuda, S., Tokimatsu, T., et~al., 2008. Kegg for
  linking genomes to life and the environment. Nucleic acids research 36~(suppl
  1), D480--D484.

\bibitem[{Keum and Nam(2017)}]{keum2017self}
Keum, J., Nam, H., 2017. Self-blm: Prediction of drug-target interactions via
  self-training svm. PloS one 12~(2), e0171839.

\bibitem[{Kingma and Ba(2014)}]{kingma2014adam}
Kingma, D.~P., Ba, J., 2014. Adam: A method for stochastic optimization. arXiv
  preprint arXiv:1412.6980.

\bibitem[{Kitchen et~al.(2004)Kitchen, Decornez, Furr, and
  Bajorath}]{kitchen2004docking}
Kitchen, D.~B., Decornez, H., Furr, J.~R., Bajorath, J., 2004. Docking and
  scoring in virtual screening for drug discovery: methods and applications.
  Nature reviews Drug discovery 3~(11), 935--949.

\bibitem[{Kuruvilla et~al.(2002)Kuruvilla, Shamji, Sternson, Hergenrother, and
  Schreiber}]{kuruvilla2002dissecting}
Kuruvilla, F.~G., Shamji, A.~F., Sternson, S.~M., Hergenrother, P.~J.,
  Schreiber, S.~L., 2002. Dissecting glucose signalling with diversity-oriented
  synthesis and small-molecule microarrays. Nature 416~(6881), 653--657.

\bibitem[{Lin et~al.(2013)Lin, Chen, and Yan}]{lin2013network}
Lin, M., Chen, Q., Yan, S., 2013. Network in network. arXiv preprint
  arXiv:1312.4400.

\bibitem[{L{\'o}pez et~al.(2017)L{\'o}pez, Dehzangi, Lal, Taherzadeh,
  Michaelson, Sattar, Tsunoda, and Sharma}]{lopez2017sucstruct}
L{\'o}pez, Y., Dehzangi, A., Lal, S.~P., Taherzadeh, G., Michaelson, J.,
  Sattar, A., Tsunoda, T., Sharma, A., 2017. Sucstruct: Prediction of
  succinylated lysine residues by using structural properties of amino acids.
  Analytical Biochemistry 527.

\bibitem[{Mousavian et~al.(2016)Mousavian, Khakabimamaghani, Kavousi, and
  Masoudi-Nejad}]{zaynabPssm}
Mousavian, Z., Khakabimamaghani, S., Kavousi, K., Masoudi-Nejad, A., 2016.
  Drug--target interaction prediction from pssm based evolutionary information.
  Journal of pharmacological and toxicological methods 78, 42--51.

\bibitem[{Mousavian and Masoudi-Nejad(2014)}]{mousavian2014drug}
Mousavian, Z., Masoudi-Nejad, A., 2014. Drug--target interaction prediction via
  chemogenomic space: learning-based methods. Expert opinion on drug metabolism
  \& toxicology 10~(9), 1273--1287.

\bibitem[{Mutowo et~al.(2016)Mutowo, Bento, Dedman, Gaulton, Hersey, Lomax, and
  Overington}]{mutowo2016drug}
Mutowo, P., Bento, A.~P., Dedman, N., Gaulton, A., Hersey, A., Lomax, J.,
  Overington, J.~P., 2016. A drug target slim: using gene ontology and gene
  ontology annotations to navigate protein-ligand target space in chembl.
  Journal of biomedical semantics 7~(1), 59.

\bibitem[{Pedregosa et~al.(2011)Pedregosa, Varoquaux, Gramfort, Michel,
  Thirion, Grisel, Blondel, Prettenhofer, Weiss, Dubourg,
  et~al.}]{pedregosa2011scikit}
Pedregosa, F., Varoquaux, G., Gramfort, A., Michel, V., Thirion, B., Grisel,
  O., Blondel, M., Prettenhofer, P., Weiss, R., Dubourg, V., et~al., 2011.
  Scikit-learn: Machine learning in python. Journal of Machine Learning
  Research 12~(Oct), 2825--2830.

\bibitem[{Rayhan et~al.(2017{\natexlab{a}})Rayhan, Ahmed, Mahbub, Jani,
  Shatabda, Farid, Rahman, et~al.}]{rayhan2017meboost}
Rayhan, F., Ahmed, S., Mahbub, A., Jani, M., Shatabda, S., Farid, D.~M.,
  Rahman, C.~M., et~al., 2017{\natexlab{a}}. Meboost: Mixing estimators with
  boosting for imbalanced data classification. arXiv preprint arXiv:1712.06658.

\bibitem[{Rayhan et~al.(2017{\natexlab{b}})Rayhan, Ahmed, Mahbub, Jani,
  Shatabda, Farid, et~al.}]{rayhan2017cusboost}
Rayhan, F., Ahmed, S., Mahbub, A., Jani, M., Shatabda, S., Farid, D.~M.,
  et~al., 2017{\natexlab{b}}. Cusboost: Cluster-based under-sampling with
  boosting for imbalanced classification. arXiv preprint arXiv:1712.04356.

\bibitem[{Rayhan et~al.(2017{\natexlab{c}})Rayhan, Ahmed, Shatabda, Farid,
  Mousavian, Dehzangi, and Rahman}]{rayhanASFMDR17}
Rayhan, F., Ahmed, S., Shatabda, S., Farid, D.~M., Mousavian, Z., Dehzangi, A.,
  Rahman, M.~S., 2017{\natexlab{c}}. idti-esboost: Identification of drug
  target interaction using evolutionary and structural features with boosting.
  Scientific reports 7~(1), 17731.

\bibitem[{Safavian and Landgrebe(1991)}]{safavian1991survey}
Safavian, S.~R., Landgrebe, D., 1991. A survey of decision tree classifier
  methodology. IEEE transactions on systems, man, and cybernetics 21~(3),
  660--674.

\bibitem[{Schomburg et~al.(2004)Schomburg, Chang, Ebeling, Gremse, Heldt, Huhn,
  and Schomburg}]{schomburg2004brenda}
Schomburg, I., Chang, A., Ebeling, C., Gremse, M., Heldt, C., Huhn, G.,
  Schomburg, D., 2004. Brenda, the enzyme database: updates and major new
  developments. Nucleic acids research 32~(suppl 1), D431--D433.

\bibitem[{Szegedy et~al.(2017)Szegedy, Ioffe, Vanhoucke, and
  Alemi}]{szegedy2017inception}
Szegedy, C., Ioffe, S., Vanhoucke, V., Alemi, A.~A., 2017. Inception-v4,
  inception-resnet and the impact of residual connections on learning. In:
  AAAI. Vol.~4. p.~12.

\bibitem[{Szegedy et~al.(2014)Szegedy, Liu, Jia, Sermanet, Reed, Anguelov,
  Erhan, Vanhoucke, and Rabinovich}]{szegedy2014going}
Szegedy, C., Liu, W., Jia, Y., Sermanet, P., Reed, S., Anguelov, D., Erhan, D.,
  Vanhoucke, V., Rabinovich, A., 2014. Going deeper with convolutions. corr
  abs/1409.4842 (2014).

\bibitem[{Szegedy et~al.(2016)Szegedy, Vanhoucke, Ioffe, Shlens, and
  Wojna}]{szegedy2016rethinking}
Szegedy, C., Vanhoucke, V., Ioffe, S., Shlens, J., Wojna, Z., 2016. Rethinking
  the inception architecture for computer vision. In: Proceedings of the IEEE
  Conference on Computer Vision and Pattern Recognition. pp. 2818--2826.

\bibitem[{Taherzadeh et~al.(2017)Taherzadeh, Zhou, Liew, and
  Yang}]{taherzadeh2017structure}
Taherzadeh, G., Zhou, Y., Liew, A. W.-C., Yang, Y., 2017. Structure-based
  prediction of protein-peptide binding regions using random forest.
  Bioinformatics, btx614.

\bibitem[{Wang et~al.(2017)Wang, You, Chen, Xia, Liu, Yan, and
  Zhou}]{wang2017computational}
Wang, L., You, Z.-H., Chen, X., Xia, S.-X., Liu, F., Yan, X., Zhou, Y., 2017.
  Computational methods for the prediction of drug-target interactions from
  drug fingerprints and protein sequences by stacked auto-encoder deep neural
  network. In: International Symposium on Bioinformatics Research and
  Applications. Springer, pp. 46--58.

\bibitem[{Wang et~al.(2013)Wang, Yang, and Li}]{wang2013drug}
Wang, W., Yang, S., Li, J., 2013. Drug target predictions based on
  heterogeneous graph inference. In: Pacific Symposium on Biocomputing. Pacific
  Symposium on Biocomputing. NIH Public Access, NIH Public Access, p.~53.

\bibitem[{Wen et~al.(2017)Wen, Zhang, Niu, Sha, Yang, Yun, and
  Lu}]{wen2017deep}
Wen, M., Zhang, Z., Niu, S., Sha, H., Yang, R., Yun, Y., Lu, H., 2017.
  Deep-learning-based drug--target interaction prediction. Journal of proteome
  research 16~(4), 1401--1409.

\bibitem[{Wishart et~al.(2008)Wishart, Knox, Guo, Cheng, Shrivastava, Tzur,
  Gautam, and Hassanali}]{wishart2008drugbank}
Wishart, D.~S., Knox, C., Guo, A.~C., Cheng, D., Shrivastava, S., Tzur, D.,
  Gautam, B., Hassanali, M., 2008. Drugbank: a knowledgebase for drugs, drug
  actions and drug targets. Nucleic acids research 36~(suppl 1), D901--D906.

\bibitem[{Xiao et~al.(2013)Xiao, Min, Wang, and Chou}]{xiao2013icdi}
Xiao, X., Min, J.-L., Wang, P., Chou, K.-C., 2013. icdi-psefpt: identify the
  channel--drug interaction in cellular networking with pseaac and molecular
  fingerprints. Journal of theoretical biology 337, 71--79.

\bibitem[{Yamanishi et~al.(2008)Yamanishi, Araki, Gutteridge, Honda, and
  Kanehisa}]{yamanishi2008prediction}
Yamanishi, Y., Araki, M., Gutteridge, A., Honda, W., Kanehisa, M., 2008.
  Prediction of drug--target interaction networks from the integration of
  chemical and genomic spaces. Bioinformatics 24~(13), i232--i240.

\bibitem[{Yamanishi et~al.(2010)Yamanishi, Kotera, Kanehisa, and
  Goto}]{yamanishi2010drug}
Yamanishi, Y., Kotera, M., Kanehisa, M., Goto, S., 2010. Drug-target
  interaction prediction from chemical, genomic and pharmacological data in an
  integrated framework. Bioinformatics 26~(12), i246--i254.

\bibitem[{Yuan et~al.(2016)Yuan, Gao, Wu, Zhang, Mamitsuka, and
  Zhu}]{yuan2016druge}
Yuan, Q., Gao, J., Wu, D., Zhang, S., Mamitsuka, H., Zhu, S., 2016. Druge-rank:
  improving drug--target interaction prediction of new candidate drugs or
  targets by ensemble learning to rank. Bioinformatics 32~(12), i18--i27.

\end{thebibliography}





\end{document}